# Learning rules from multisource data for cardiac monitoring

## Elisa Fromont*

UMR CNRS 5516, Laboratoire Hubert Curien,
Université de Lyon, Université de Saint Etienne,
 18 Rue du Professeur Benoît Lauras
42000 Saint-Etienne, France
E-mail: elisa.fromont@univ-st-etienne.fr
*Corresponding author

## René Quiniou

IRISA, Campus de Beaulieu,
F-35042 Rennes Cedex, France
E-mail: quiniou@irisa.fr

## Marie-Odile Cordier

IRISA, Campus de Beaulieu,
F-35042 Rennes Cedex, France
E-mail: cordier@irisa.fr

**Abstract:** This paper formalises the concept of learning symbolic rules from multisource data in a cardiac monitoring context. Our sources, electrocardiograms and arterial blood pressure measures, describe cardiac behaviours from different viewpoints. To learn interpretable rules, we use an Inductive Logic Programming (ILP) method. We develop an original strategy to cope with the dimensionality issues caused by using this ILP technique on a rich multisource language. The results show that our method greatly improves the feasibility and the efficiency of the process while staying accurate. They also confirm the benefits of using multiple sources to improve the diagnosis of cardiac arrhythmias.

René Quiniou has been a Research Associate at INRIA Rennes-Bretagne Atlantique Research Center, France since 1981. His main research interests are machine learning and data mining methods aiming at discovering dynamic models for future use in model-based diagnosis and monitoring.

Marie-Odile Cordier has been a Full Professor at the University of Rennes 1, France, since 1988. In the past ten years, she has focused her research on the diagnosis of complex systems.

## 1  Introduction

Monitoring devices in Cardiac Intensive Care Units (CICU) use only data from Electrocardiogram (ECG) channels to automatically diagnose cardiac arrhythmias. However, data from other sources like arterial pressure, phonocardiograms, ventilation, etc., are often available. This additional information could be used in order to improve the diagnosis and, consequently, to reduce the number of false alarms emitted by monitoring devices. From a practical point of view, only severe arrhythmias (considered as *red alarms*) are diagnosed automatically, and in a conservative manner, to avoid missing a problem. The aim of the work that has begun with the CALICOT project (Carrault et al., 2003) is to improve the diagnosis of cardiac rhythm disorders in a monitoring context and to extend the set of recognised arrhythmias to non-lethal ones if they are detected early enough (considered as *orange alarms*). To achieve this goal, we combine information coming from several sources, such as ECG and Arterial Blood Pressure (ABP) channels as shown in Figure 1.

**Figure 1**   Multisource data: lead I and V of an electrocardiogram and an arterial blood pressure channel

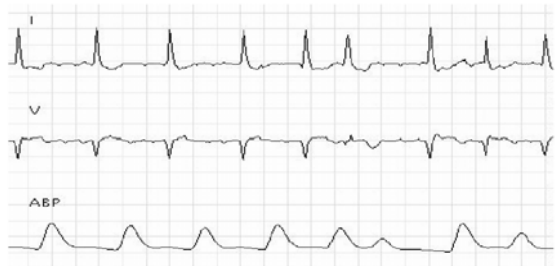

We are particularly interested in learning temporal rules that enable such a multisource detection scheme. To learn these kinds of rules, a relational learning system based on ILP is well-adapted. ILP not only enables to learn relations between characteristic events occurring on the different channels but also provides rules that are understandable by doctors, since the representation method relies on first order logic.

One possible way to combine information coming from difference sources is to aggregate all the learning data and then, to learn as in the monosource (i.e., one data source) case. However, in a multisource learning problem, the amount of data and the expressiveness of the language increase dramatically and with them, the computation time of ILP algorithms and the size of the hypothesis search space. Many methods have been proposed in ILP to cope with the search space dimensions; one of them is using a declarative bias (Nédellec et al., 1996). This bias aims either at narrowing the search space or at ranking hypotheses to consider first the better ones for a given problem. Designing an efficient bias for a multisource problem is a difficult task. In Fromont et al. (2004), we have sketched a divide-and-conquer strategy (called biased multisource learning) where symbolic rules are learned independently from each source and then, the learned rules are used to bias automatically and efficiently a new learning process on the aggregated dataset. This proposal is developed here and applied on cardiac monitoring data.

In the Section 2, we introduce the vocabulary and the medical background needed to understand the examples throughout the paper. In the Section 3, we give a brief introduction to ILP and we extend this formalism to multisource learning. In the Section 4 we describe our method to learn multisource rules in an efficient way. In the Section 5, we describe our experiments to compare the performances of ILP on cardiac data when using a single source, multiple sources with a naive method and multiple sources with the method described in the previous section. Section 6 is dedicated to related work. The last section gives conclusions and perspectives.

## 2 Background on the cardiological application

The contraction of the cardiac muscle is due to the propagation of an electric wave that spontaneously starts in a precise location of the heart. An *Electrocardiogram* (ECG) is a graphical representation of this electrical activity, measured by placing different electrodes at specific points on the patient's body. An ECG is made of several *leads*. A lead (cf. leads I and V of Figure 1) records the electrical signals of the heart from a particular combination of recording electrodes.

The electrical propagation represented on the ECG, can be described by a succession of characteristic waves. In cardiology, those waves are labelled by letters starting from P. The most informative of those waves are the *P wave*, and the succession of the Q, R and S waves called the *QRS complex*. Figure 2 shows a normal cardiac cycle (the figure omits other waves that are not useful for the comprehension of the paper).

**Figure 2** Characteristic waves of an ECG : a normal cycle (see online version for colours)

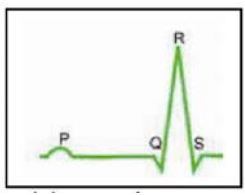

A *cardiac arrhythmia* occurs when the electrical activity becomes abnormal. As a consequence, the heart rhythm can become faster, slower or irregular and the path that takes the electric waves through the heart can change. This can be seen on the ECG by noticing longer/shorter intervals between the characteristic waves, missing waves or different wave shapes.

The *Arterial Blood Pressure* (ABP) depends on the contraction of the cardiac muscle (and therefore indirectly, also on the propagation of the electrical wave). The heart pumps the blood through the vessels such as the arteries. ABP measures the blood pressure in the larger arteries. The most informative points in an ABP curve are the instants when the pressure is the lowest (the *diastole*) and when the pressure is the highest (the *systole*).

## 3    Multisource learning with Inductive logic programming

In this section, we give a brief introduction to ILP (see Muggleton and De Raedt, 1994, for more details) and we give a formalisation of this paradigm applied to multisource learning. We assume familiarity with first order logic (see Lloyd, 1987, for an introduction).

### 3.1    Introduction to ILP

Inductive Logic Programming (ILP) is a supervised symbolic machine learning method. Given a set of examples `E` and a set of general rules `B` representing the background knowledge, it builds a set of hypotheses `H`, in the form of classification rules for a set of classes `C`. `B` and `H` are logic programs i.e., sets of rules (also called definite clauses) having the form `h:-b₁, b₂, ..., bₙ`. This rule can be read "if `b₁` and `b₂` and ... and `bₙ` are true then `h` is true". When `n = 0`, i.e., there is no condition to make `h` true, such a rule is called a fact. `E` is a labelled set of ground facts. *A multi-class ILP problem* is an ILP problem which involves multiple classes. In our cardiological example, we want to learn hypotheses that describe multiple arrhythmias. In a multi-class problem, each example `e` of `E` labelled by `c` is a *positive example* for the class `c` and a *negative example* for the class `c' ∈ {C-c}`. In our cardiological example, examples are descriptions of signal that are known to be associated with a particular arrhythmia. Negative examples for this particular arrhythmia are all examples associated to another arrhythmia.

The following definition for a multi-class ILP problem is inspired by (Blockeel et al., 1999).

**Definition 1:** A multi-class ILP problem is described by a tuple
`<L, E, B, C>` such that:

- `E = {(eₖ, c) | k ∈ [1, m], c ∈ C}` is a set of m examples where each $e_k$ is a set of facts expressed in the language `L_E`.

- *B* is a set of rules expressed in the language $L = L_E \cup L_H$ where $L_H$ is the language of hypotheses.

The ILP algorithm has to find a set of rules $H = \{H_c \mid c \in C\}$ such that for each $(e, c) \in E$:

1    $H_c \wedge e \wedge B \models c$

2    $\forall c' \in \{C - c\}, H_c \wedge e \wedge B \not\models c'$.

The first condition of Definition 1 means that the class of an example must be deducible from the description of the example, the background knowledge `B` and the set of hypotheses `H`$_c$. In this case, we say that `H`$_c$ covers `(e, c)` written `covers(H`$_c$`, (e,c))`. The second point means that hypotheses must be discriminative i.e., an hypothesis `H`$_c$ for a class `c` should not cover an example of another class.

The hypotheses in `H`$_c$ for each class *c* are searched for in a so-called *hypothesis space*. A generalisation relation, usually the *θ*-subsumption (Plotkin, 1970), can be defined on hypotheses. This relation induces a lattice structure on $L_H$ which enables a more efficient exploration of the search space. Many ILP systems have been developed over the years such as ICL[1] (De Raedt and Van Laer, 1995), or Aleph (Srinivasan, 2003) and its predecessors Progol (Muggleton, 1995), Foil (Quinlan and Cameron-Jones, 1993), Tilde (Blockeel and De Raedt, 1998), etc. Those systems differ by their semantics (Definition 1 in the case of ICL), their tools to limit the hypothesis search space (called *bias*) and the strategy they use to explore this space.

For example, ICL explores the search space from the most general clause to more specific clauses and looks for hypotheses (for each class) that fulfil conditions 1 and 2 of Definition 1. The search stops when a clause that covers no negative example while covering some positive examples is reached. At each step in the exploration of the search space, the best clause so far according to some criteria (for e.g., accuracy) is refined by adding new literals to its body, applying variable substitutions, etc.

For the work presented in this paper, we have specifically decided to use ICL because it uniquely provides a tool called *DLAB* (De Raedt and Dehaspe, 1997) to restrict the search space, initially defined by $L_H$ and makes the search space a lot more efficient in presence of background information on the domain. *DLAB* is a declarative language bias which allows us to define, syntactically, the sub-set of clauses from $L_H$ which belong to the search space. More precisely, a DLAB grammar defines exactly which literals are allowed in hypotheses, in which order literals are added to hypotheses during the search but also the search depth limit (the clause size).

One can specify a DLAB grammar with expressions such as `l-h : [el1, el2, ... ,eln]` which means "choose between `l-h` elements in the list `[el1, el2, ..., eln]`". The `len` symbol is used to specify the total length of the list. For example, the DLAB term `p(2-len :[el2, el3])` can be used to generate the literals `p(el1, el2)`, `p(el1, el3)`, `p(el2, el3)`or `p(el1, el2, el3)`. An example of such grammar is given in Figure 3. It specifies that an hypothesis representing a cardiac beat sequence has the following structure:

- A first cardiac beat composed of a P wave called `P1` followed by a QRS complex called `R1` followed by some optional predicates (constraint `0-len` line 7) `pr` and `rr` lines 7–8. Attributes with values *normal* or *abnormal* are associated with P waves and QRS complexes. The predicates `pp`, `pr` or `rr` specify the delay between two P waves, a P wave and a QRS or two QRSs, respectively. A delay can be *short*, *normal* or *long*.

- The first beat can be followed by a second optional beat (the block that starts by `0-len in` line 7 is closed at the end of line 13). Line 8 states that, in such a case, a P wave called P2, associated with a mandatory predicate `pp` (line 10) immediately (there cannot be another event in-between) follows the QRS complex `R1` (line 9).

- Line 11 specifies another optional QRS (`R2`) associated to an optional predicate `rr` (line 13) which follows (not necessarily immediately) `R1`.

**Figure 3**   Syntactic description of a cardiac cycle using DLAB

```
1      len-len:[
2          p(P1,1-1:[normal,abnormal]),
3          suc(P1,R0),qrs(R1,1-1:[normal,abnormal]),
4          suc(R1,P1),
5          0-len:[rr(R0,R1,1-1:[short,normal,long]),
6          pr(P1,R1,1-1:[short,normal,long])]],
7          0-len:[
8          len-len:[p(P2,1-1:[normal,abnormal]),
9          suci(P2,R1),
10         pp(P1,P2,1-1:[short,normal,long])]],
11         len-len:[qrs(R2,1-1:[normal,abnormal]),
12         suc(R2,R1),
13         0-1:[rr(R1,R2,1-1:[short,normal,long])]]]]
14     ],
```

For example, the two following clauses can be induced from the DLAB grammar described in Figure 3:

```
class(x):-  (this part is not described in the grammar)
p(P1,normal), suc(Ro,P1)
qrs(R1,abnormal), suc(R1,P1),
pr(RO,R1,short).
```

If ICL learns this clause, an arrhythmia of type x can be diagnosed if, at some point in the signal, there is a normal P wave, followed by an abnormal QRS complex such that the interval between the P wave and the QRS complex is short.

```
class(y):-
p(P1, normal), suc(P1, RO), qrs(R1, normal),
pr(P0, R0, long), suc(R1, P1),
p(P2, anormal),suci(P2, R1),
pp(P1, P2,short),
qrs(R2, abnormal), suc(R2, R1).
```

If ICL learns this clause, an arrhythmia of type y can be diagnosed if there is a normal P wave on the signal, followed by a normal QRS complex such that the interval between the P wave and the QRS complex is long; another P wave should immediately follow this QRS such that the interval between this P wave and the first one is short. The last P wave (P2), should be followed by an abnormal QRS.

In the following, the most specific clauses of the search space that can be generated from a bias specification will be called *bottom clauses* in reference to the bottom clauses used in some ILP systems such as *Progol* (Muggleton, 1995) or *Aleph* (Srinivasan, 2003). We call *event predicates* (resp. *literals*), predicates (resp. literals) that describe events occurring in some data source, e.g., *qrs* (resp. *qrs(R0,normal)*). *Relational literals* are literals that associate several events occurring on the same or on different sources, e.g., *suc(R0,R1)* or *rr1(R0,R1,normal)*. *Global relational predicates* are relational predicates common to the languages used to describe each sources, e.g., *suc*. Those predicates can be used to associate events occurring in different sources.

### 3.2 Multisource learning

In a multisource learning problem, examples are bi-dimensional; the first dimension, $i \in$ `[1, s]`, refers to a source, the second one, $k \in$ `[1, m]`, refers to a situation. Examples indexed by the same situation correspond to contemporaneous views of the same phenomenon.

Aggregation is the operation consisting of merging examples from different views of the same situations. The aggregation function $F_{agg}$ depends on the learning data type and can be different from one multisource learning problem to another. Here, the aggregation function is simply the set union associated to inconsistency elimination. Examples are considered inconsistent on the different sources if $\exists i, j, k, i \neq j, (e_{i,k}, c) \wedge (e_{j,k}, c') \wedge c \neq c'$. This definition of inconsistency is close that of Blum and Mitchell (1998), used for compatible function in the context of co-training. In this paper, we consider that inconsistent examples have been removed from the database before starting the multisource learning process. Moreover, the user might have some knowledge about how to combine the different sources (temporal constraints between events occurring on each sources, associated events, …). This knowledge, called the aggregation knowledge, is described in the background knowledge `B`.

Multisource learning for a multi-class ILP problem is then defined as follows:

**Definition 2** (Multisource learning with ILP)**:** Let $i \in$ `[1, s]` be the number of sources. Let $<$ `L_i, E_i, B_i, C` $>$ be ILP problems such that:

- $E_i =$ `{ (e_{i,k}, c) | k ∈ [1, p], c ∈ C}` is the set of p consistent examples for each source $i$.

- $B_i$ (background knowledge) is the set of rules expressed in the language $L_i$ for each source $i$.

A multisource ILP problem is defined by a tuple $<L, E, B, C>$ such that:

- $E = F_{agg}(E_1, E_2, ..., E_s) =$ `{ (e_k, c) | e_k = ` $\bigcup_{i=1}^{s} e_{i,k}$, `k ∈ [1, p]}`

- $L = L_E \cup L_H$ is the multisource language

where: $L_E = F_{agg}(L_{E1}, L_{E2}, ..., L_{Es})$ and

$\bigcup_{i=1}^{s} L_{Hi} \subseteq L_H$

- $B$ is a set of rules in the language $L$.

The ILP algorithm has to find a set of rules $H =$ `{H_c | c ∈ C}` such that for each `(e, c) ∈ ` $E$:

1   $H_c \wedge e \wedge B \models c$

2     $\forall c' \in \{C - c\}, H_c \wedge e \wedge B \mid \neq c'.$

A *naive* multisource approach consists of learning directly from the aggregated examples and with a global bias that covers the whole search space related to the aggregated language `L`. In this case, the hypothesis language `L_H` is the product of the languages used to describe all the different sources i.e., the union of all the languages `L_i` (all event predicates and relational predicates used to describe each source i) plus all relationships that can exist between events occurring on the different sources. The main drawback of this approach is the size of the resulting search space. In many situations, the learning algorithm is not able to cope with it or takes too much computation time. The only solution is to specify an efficient language bias, but this is often a difficult task especially when no expert information describing the relations between sources is provided. In the following section, we propose a new method to automatically create such a bias.

## 4     Reducing the multisource learning search space

In this section, we describe a two-steps strategy, called *biased multisource learning*, to efficiently learn multisource rules using ILP. First, we propose to learn rules independently from each source (this step is called *monosource learning*). The resulting clauses are then combined to build a DLAB bias that will be used in a second step, for a new multisource learning process on the aggregated datasets. In our application in cardiology, the second step can be seen as a method to synchronise the monosource rules learned during the first step. Synchronisation of events occurring on the different sources is described by a predicate *suci(X, Y)* (for *immediate succession*). This predicate means that event *X* immediately follows event *Y*. Algorithm 1 shows the different steps of the method on two source learning. It can be straightforwardly extended to *n* source learning. We assume that the situations are described using a common reference time. This is seldom the case for raw data, so we assume that the data sets have been preprocessed to ensure this property.

### Algorithm 1

1     **Learn** with bias *Bias₁* on the multi-class ILP problem <$L_1, E_1, B_1, C$>. Let $H_{c1}$ be the set of rules learned for a given class $c \in C$.

2     **Learn** with bias *Bias₂* on the multi-class ILP problem <$L_2, E_2, B_2, C$>. Let $H_{c2}$ be the set of rules learned for the class c.

3     **Aggregate** the sets of examples $E_1$ and $E_2$ giving $E_3$.

4     **Generate** for all c and from all pairs $(h_{1j}, h_{2k}) \in H_{c1} \times H_{c2}$ a set of bottom clauses $BT_c$ such that each $bt_i \in BT_c$ built from $h_{1j}$ and $h_{2k}$ is more specific than both $h_{1j}$ and $h_{2k}$. The literals of $bt_i$ are all the literals of $h_{1j}$ and $h_{2k}$ plus new relational literals (suci) that synchronise events in $h_{1j}$ and $h_{2k}$.

5     For each $c \in C$, **Build** bias $Bias_c$ from $BT_c$. Let $Bias_3 = Bias_c \mid c \in C\}$.

6     **Learn** with $Bias_3$ on the problem <$L, E_3, B_3, C$> where :

•     $L$ is the multisource language (as defined in Section 3.2)

- $B_3$ is a set of rules expressed in the language *L*.

## 4.1 Generation of the bottom clauses

One goal of multisource learning is to make relationships between events occurring on different sources explicit. For each pair of hypotheses $(h_{1j}, h_{2k})$ learned on each source separately, there are as many bottom clauses as ways to intertwine events from the two sources. The *suci* predicates are only introduced if two events occurring on different sources may be synchronised. In the clause $bt_n$ of Figure 4 the relational predicate suc(R0, P0) means that the wave R0 occurs after the wave P0 on the first source (ECG). However, some events may occur between P0 and R0 on other sources. The predicate suci(D0, R0) is more precise: it means that no event from any source can occur between D0 and R0.

**Figure 4**   Example of bottom clauses that may be generated from the pair of clauses (h₁, h₂) describing the same class x

```
Let h₁ =
class(x):-   %sequence P0-R0
p(P0,normal), qrs(R0,normal),
pr1(P0,R0,normal),  suc(R0,P0).

be the rule induced for class x on the data from source 1.

Let h₂ =
class(x)   :-   %sequence D0-S0
diastole(D0,normal),
systole(S0,normal), suc(S0,D0).

be the rule induced for the same class x on the data from source 2.
The generated bottom clauses for the pair (h₁,h₂) are :

bt₁ =
class(x)   :-   %sequence P0-D0-R0-S0
p(P0,normal), diastole(D0,normal),
suci(D0,P0),qrs(R0,normal), pr1(P0,R0,normal), suci(R0,D0),
suc(R0,P0),systole(S0,normal), suci(S0,R0), suc(S0,D0).

bt₂ =
class(x)   :-   %sequence D0-P0-S0-R0
diastole(D0,normal), p(P0,normal),
suci(P0,D0),systole(S0,normal), suci(S0,P0), suc(S0,D0),
qrs(R0,normal), pr1(P0,R0,normal),
suci(R0,S0), suc(R0,P0).
...,
bt₃ =
class(x)   :-   %sequence D0-S0-P0-R0
diastole(D0,normal), systole(S0,normal),
suc(S0,D0), p(P0,normal),
suci(P0,S0), qrs(R0,normal),
pr1(P0,R0,normal), suc(R0,P0).
...,
btₙ =
class(x)   :-   %sequence P0-R0-D0-S0
p(P0,normal), qrs(R0,normal),
pr1(P0,R0,normal), suc(R0,P0),
diastole(D0,normal), suci(D0,R0),
systole(S0,normal), suc(S0,D0).
```

The number of bottom clauses that can be generated for one pair $(h_{1j}, h_{2k})$, is $C^n_{n+p}$, where $n$ is the number of event predicates belonging to $h_{1j}$ and $p$ is the number of event predicates belonging to $h_{2k}$. This number can be very high if there are many event predicates in each rule. However, in practice, many bottom clauses are not generated because the related event sequences do not make sense for the application. For example, in our application in cardiology, an expert would forbid every sequence where an event from source 1 (ECG) occurs between the events *diastole(A, B)* and *systole(C, D)* occurring on source 2 (ABP). In the example in Figure 4, the expert would then remove $bt_1$ and $bt_2$ from the set of valid bottom clauses. The number of clauses in $BT$ is the total number of valid bottom clauses generated from all possible pairs. The bias can then be generated automatically from this set of bottom clauses.

### 4.2   *Automated bias construction from a set of bottom clauses*

Each previously constructed bottom clause defines a search space. The biased multisource search space consists in the concatenation of all search spaces defined by the bottom clauses. In each such search space, hypotheses looked for must be equal or more general than the bottom clause that bounds this search space. Each bottom clause can be seen as a pool of literals that can be added to an hypothesis when exploring the search space. To make the search spaces finite, some constraints are added on hypotheses:

1   for a given search space, the number of literals in a hypothesis is lower than, or equal to, the number of literals in the bottom clause that bounds the search space

2   a relational literal can be added in any hypothesis only if it relates events also appearing in the hypothesis in the form of event literals

3   any event appearing in an event literal must be related to some other event by a global relational predicate.

This construction ensures that each node in the search space corresponds to a semantically valid hypothesis. Indeed, impossible sequences have been eliminated during the generation of the bottom clauses, and literals used in the hypothesis for one class are strongly connected to the corresponding class since they already appear in a monosource rule learned for this class.

From this set of constraints, we construct a DLAB bias such that the search space restricted by the bias is exactly the 'union' of all the search spaces previously defined. This is simply done by defining a DLAB expression `1-1:[...]` between all the DLAB blocks constructed from the bottom clauses. Figure 5 shows the different search spaces considered during a biased multisource learning step on two sources. *L* is the naive multisource search space corresponding to the product of all the monosource learning languages. $L_1$ et $L_2$ are the monosource search spaces. The dashed area is the union of all the search spaces defined by the bottom clauses `bt_1, ..., bt_n`. The search space associated with one bottom clause corresponds to a white line. This space contains all hypotheses that are more general than the bottom clause and that respect the constraints previously defined.

Each search space bounded by a bottom clause is represented by a DLAB block. Each block allows to induce an hypothesis equal to, or more general than, the considered bottom clause. Figure 6 shows a part of the DLAB bias corresponding to the bottom clause bt$_n$ from Figure 4. All the search spaces (blocks) can be explored to find the best multisource hypothesis but only one block is sufficient to learn such an hypothesis.

**Figure 5** Construction of the biased multisource search space (see online version for colours)

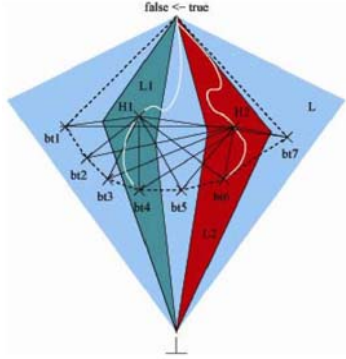

**Figure 6** Example of a DLAB bias constructed from a set of bottom clauses

```
1-1:[ one of the following block is chosen :
...,
1-1:[len-len:[...]], (n-1)th bloc : bt_{n-1}
1-1:[len-len:[ bt_n
 p(P0,normal),qrs(R0,normal),suc(R0,P0),
  0-1:[pr1(P0,R0,normal)]
 ],
  0-1:[len-len:[
  len-len:[
  diastole(D0,normal),suci(D0,R0)
 ],
  0-1:[len-len:[
  len-len:[
  systole(S0,normal),suc(S0,D0)
 ]
]]]]]
]
```

## 4.3 Properties of the biased multisource search space

The construction of the multisource search space and the way examples from each source are aggregated confer Properties 2 and 3 to the multisource search space.

**Property 1 (Aggregated examples):** *Let s be the number of sources and $F_{agg}$ the set union. Let $L_i ev$ be the language $L_i$ restricted to event predicates. Let $H_{ic}$ be an hypothesis learned on source i describing class c. Let $c' \in C$ such that $c' \neq c$.*

1  *covers($H_{ic}$,($e_{i,k}$,c)) $\Rightarrow$ covers($H_{ic}$,($e_k$,c)).*

2  *(($L_i ev \cap \bigcup_{j=1, j \neq i} L_j ev = \varnothing) \wedge$ (covers($H_{ic}$,($e_{i,k}$,c'))))*

    *$\Rightarrow \neg covers(H_{ic}$,($e_k$,c')).*

The first point of Property 1 means that if the learned hypothesis ($H_{ic}$) covers the *monosource* example, ($e_{i,k}$,c) then $H_{ic}$ also covers the aggregated example *($e_k$,c)*. The second point means that if the languages $L_i$ that describe each source have no common event predicate and if the learned hypothesis $H_{ic}$ does not cover example ($e_k$, $c'$) then $H_{ic}$ does not cover the aggregated example ($e_{i,k}$, $c'$). Proof 1 demonstrates this proposition.

*Proof 1*:

1   Examples coming from the different sources are consistent so, if there exists $i = 1,s$ such that $H_{ic} \wedge e_{i,k} \wedge B \mid = c$ then $H_{ic} \wedge [e_{1,k}, e_{2,k}, \ldots, e_{n,k}] \wedge B \mid = c$ and so $H_{ic} \wedge e_k \wedge B \mid = c$.

2   $((L_i ve \cap \bigcup^s_{j=1, j \neq i} L_j ev = \varnothing) \Rightarrow \forall k,\ \forall j \neq i,\ \forall c' \in C-\{c\},\ \neg\text{covers}(H_{ic}, (e_{j,k}, c'))$.

    $\forall k,\ \forall c' \in C-\{c\},\ \neg\text{covers}(H_{ic}, (e_{i,k}, c'))$.

So $\forall k,\ \forall l,\ H_{ic} \wedge l_{1,k} \wedge B \mid \neq c$.

Moreover, examples coming form different sources are consistent so $H_{ic} \wedge [e_{1,k}, e_{2,k}, \ldots, e_{n,k}] \wedge B \mid \neq c'$. In conclusion $H_{ic} \wedge _k \wedge B \mid \neq c'$ and so $\neg\text{covers}(H_{ic}, (e_k, c'))$.

**Property 2** (Correctness)**:** There exists hypotheses with an equal or higher training accuracy than the training accuracy of $H_{c1}$ and $H_{c2}$ in the search space defined by $\text{Bias}_3$ of Algorithm 1.

The accuracy[1] is defined as the rate of correctly classified examples. Because of Property 1, the hypotheses from $H_{c1}$ and $H_{c2}$ can also be learned in the second step of the biased multisource learning algorithm if the biased multisource search space does not contain any better solutions. However, the multisource search space restricted by the bias $Bias_3$ constructed in point 5 of Algorithm 1 does not necessarily contain the best multisource solutions for the multisource problem. Indeed, if we consider the temporal window which contains all events occurring in a monosource rule, it is not possible to learn composite multisource rules with the biased multisource method if there is no overlapping between the temporal windows of the monosource rules.

**Property 3** (Search space reduction)**:** The search space defined by $\text{Bias}_3$ in Algorithm 1 is smaller than the multisource naive search space.

The size of the search space specified by a DLAB bias can be computed by the method given in De Raedt and Dehaspe (1997). The biased multisource search space is smaller than the naive search space since the language used in the first case is a subset of the language used in the second case.

In the next section, this biased multisource method is compared to monosource learning from cardiac data coming from an ECG for the first source and from measures of ABP for the second source. The method is also compared to a naive multisource learning performed on the data aggregated from the two former sources.

## 5   Evaluation and experimental results

### 5.1   Data

We use the Multi-parameter Intelligent Monitoring for Intensive Care database (MIMIC) (Moody and Mark, 1996) which contains 72 patients' files recorded in the CICU of the Beth Israel Hospital Arrhythmia Laboratory. Raw data concerning the channel V1 of an ECG and an ABP signal are extracted from the MIMIC database and transformed into symbolic descriptions (our examples) such as the one given in Figure 7. To create our ICL examples, we first use QRS, P, diastole and systole detectors (Carrault et al., 2003; Portet, 2008; Hoeksel et al., 1997) to process the raw data. We then obtain a first labelling in terms of instant amplitude of the pressure for the ABP channel and shape for the ECG waves. This labelling is verified and completed by experts to ensure the quality of the data. The labels are then automatically transformed into a logical knowledge database (Prolog facts) suitable for ICL.

**Figure 7**   Representation of an example in ICL: a ventricular doublet described on the ECG (left) and on the pressure channel (right) in ICL.

```
begin(model).              begin(model).
doublet_3_I.               rs_3_ABP.
.....                      .....
p(p7,4905,normal).         dias(pd4,3406,80).
qrs(r7,5026,normal).       suc(pd4,ps3).
suc(r7,p7).                sys(ps4,3558,120).
qrs(r8,5638,abnormal).     suc(ps4,pd4).
suc(r8,r7).                 ......
qrs(r9,6448,abnormal).     end(model).
suc(r9,r8).
.....
end(model).
```

Figure 7 (left) shows seven facts describing a ventricular doublet example, one P wave occurring at time 5026 and three QRS complexes as well as relations stating the order in which these waves occur in the sequence. Additional information such as the wave shapes (normal/abnormal) is also provided in the predicate. Figure 7 (right) provides a similar description for the pressure channel. The last argument of the predicates `dias` and `sys` is the amplitude of the blood pressure in millimetres of mercury. Seven cardiac rhythms (corresponding to seven classes) are investigated in this work: normal/sinus rhythm (*sr*), ventricular extra-systole (*ves*), bigeminy (*bige*), ventricular doublet (*doublet*), ventricular tachycardia (*vt*) which is considered as being a red alarm in CICU, supra-ventricular tachycardia (*svt*) and atrial fibrillation (*af*).

### 5.2   Experimental process

To verify empirically that the biased multisource learning method is efficient, we have performed three kinds of learning experiments on the same learning data: monosource learning from each source, naive multisource learning from aggregated data using a bias as minimally restrictive as possible (allows every possible sequence of events predicates with all possible relations between them) and multisource learning using a bias constructed from the rules discovered by monosource learning in the first experiments. In order to assess the impact of the learning hardness, we have performed three series of experiments: in the first series (5.3) the representation language was expressive enough to give good results for each setting; in the second series (5.4) the expressiveness of the representation language was reduced drastically. In the last experiment (5.5), we have used complementary data to emphasise the advantage of using multiple sources to ensure a reliable diagnosis. Three criteria are used to compare the learning results: computational load (CPU time in second), accuracy and complexity (*Comp*) of the rules (each number in a cell represents the number of cardiac cycles in each rule produced by the ILP system). As the number of examples is rather low, a *leave-one-out* cross validation method is used to assess the different criteria. The average accuracy values obtained during cross-validation training (*TrAcc*) and test (*Acc*) are provided. The leave-one-out (i.e., cross-validation with number of folds = number of examples) technique for the biased multisource learning is detailed in Algorithm 2.

**Algorithm 2** (Cross-validation for biased multisource learning)**:**

*Let $E = \{(e_k, c)|e_k = \bigcup_{i=1}^{s} e_{i,k}, \ k \in [1, \ n]\}$ be the set of aggregated multisource examples. Let $p$ be the number of folds in the cross-validation. For a given class c:*

- *For each source i, perform a p-fold cross-validation to learn p theories. At each cross-validation step, n/p consistent examples are removed from the learning database and are kept for the test. For each source, always remove the same set of examples in the same order $T_{i,1}, T_{i,2}, \ldots, T_{i,p}$.*

- *Construct p multisource biases using the p theories learned for each source.*

- *Create p sets of multisource aggregated examples $E_p$ such that $E_p = E − T_p$. The removed examples can be either positive negative examples for class c.*

- *Do p biased multisource learning. TrAcc is the average accuracy of each learning step (TP + TN)/(TP + TN + TP + TN). Acc is the average test accuracy obtained when testing the p removed examples on the p learned multisource theories. If a removed example is $(e_p, c)$, the test equal 1 if the example is an interpretation of the theory, 0 otherwise; if a removed example is $(e_p, c')$, $c \neq c'$, the test equal 1 if the removed example is contradictory with the learned theory, 0 otherwise.*

The learning settings and the multisource background knowledge $B$ do not change from one cross validation fold to another. To keep good properties within the biased multisource search space, the settings of the ILP system used must be the same during the monosource learning steps and the biased multisource learning step. The simplest way to create the biased multisource background knowledge is to define $B$ as the union of all $B_i$ and simply add the definition of the *suci* predicate

### 5.3   Learning from the whole database

Tables 1 and 2 give an idea of the computational complexity of each learning method (monosource on ECG and ABP channels, naive and biased multisource on aggregated data) for the ILP system. *Nodes* is the number of nodes explored in the search space and *Time* is the learning computation time in CPU seconds on a Sun Ultra-Sparc 5. If the number of nodes that must be evaluated during the search is too high (the search space is too big) it is more likely that the search will end up in a sub-optimal part of the space where no good solutions can be found even if one actually exists (it can also happen for smaller search space but it is less likely). The ILP system can also run out of memory if the amount of information needed to be stored in order to try different possibilities during the search is too large. As explained in Section 3, the size of the search space depends on the richness of the language (the number of predicates that can be used) and on the number of possible substitutions (related to the number of possible values for each argument of the predicates) that can be done.

**Table 1** Size of the the search space (number of visited nodes = number of refinements) for each class and learning computation times (in second).

|  | Monosource: ECG | | Monosource: ABP | |
|---|---|---|---|---|
|  | *Nodes* | *Time* | *Nodes* | *Time* |
| Sr | 4634 | 3660 | 5475 | 1415 |
| Ves | 1654 | 189 | 19971 | 2362 |
| Bige | 708 | 160 | 7365 | 337 |
| Doublet | 790 | 125 | 9840 | 596 |
| Vt | 428 | 109 | 10833 | 691 |
| Svt | 232 | 105 | 1326 | 438 |
| Af | 64 | 4 | 6477 | 1923 |

**Table 2** Size of the the search space (number of visited nodes) for each class and learning computation times (in second). for naive and multisource learning. Time* includes the monosource learning computation times

|  | Naive multisource | | Biased multisource | | |
|---|---|---|---|---|---|
|  | *Nodes* | *Time* | *Nodes* | *Time* | *Time** |
| Sr | 12889 | 36984 | 415 | 471 | 5546 |
| Ves | 12887 | 22347 | 2020 | 951 | 3502 |
| Bige | 15103 | 22298 | 77 | 23 | 520 |
| Doublet | 19084 | 7832 | 346 | 105 | 826 |
| Vt | 4317 | 7046 | 140 | 205 | 1005 |
| Svt | 2155 | 14544 | 75 | 114 | 657 |
| Af | 135 | 325 | 16 | 47 | 1974 |

Tables 1 and 2 show that the number of visited nodes during the learning process is smaller for the biased multisource method than for the lowest monosource learning process (here, ECG). This confirms that the search space created using the two (for a two sources problem) monosource rules is smaller than the 'union' of both monosource search spaces. Besides, on average, 50 times less nodes are explored during biased multisource learning than during naive multisource learning, which confirms that our method allows a huge reduction of the complexity of the multisource problem. However, the computation time does not grow linearly with the number of explored nodes because the covering tests (determining whether an hypothesis is consistent with the examples) are more complex for multisource learning. Biased multisource learning computation times take into account monosource learning computation times and is still very much shorter than for naive multisource learning (on average, 13 times lower).

Tables 3 and 4 give the average accuracy and complexity of rules obtained during cross validation for the monosource and the two multisource learning methods. The accuracy of the monosource rules is very good for ECGs where a maximal learning accuracy (1) is obtained for every class except sinus rhythm (*sr*) and ventricular extra-systole (*ves*). However, perfect results are also obtained for these two last classes when using the ABP source. Such results give no hope of better results by using both sources together, especially with the biased multisource method. For this experiment, the biased multisource method behaves as a voting method and selects for each class the monosource rule that has the best training accuracy. Thus, the biased multisource rules are equal to the ECG monosource rules for arrhythmias *bige*, *doublet*, *vt*, *svt* and *af* and equal to the ABP monosource rule for *sr*. According to the theoretical properties of the biased multisource search space, the training learning accuracy of the biased multisource learned rules is better than, or equal to, the learning accuracy of the best monosource rule. The biased multisource rules learned for *ves* takes into account sometimes only events from the ECG and sometimes only events from the ABP which results a fair test accuracy (*Acc*) for this class. We can also notice that, for this experiment, biased multisource learning accuracy is always better than or equal to the naive multisource learning accuracy.

The rules learned for *svt* in the four learning settings are given in Figures 8–11. All those rules are perfectly accurate. The predicate *cycle_abp(D, ampsd, S, ampds)* is a kind of macro predicate that expresses the succession of a diastole named *D*, and a systole named *S*. *ampsd* (resp. *ampds*) expresses the symbolic pressure variation ($\in \{low$, *normal*, *high*$\}$) between a systole and the following diastole *D* (resp. between the diastole *D* and the following systole *S*). The biased multisource rule and the naive multisource rule are very similar but specify different event orders (in the first one the diastole-systole specification occurs before two close-in-time QRS, whereas in the second one, the same specification occurs after two close-in-time QRS).

**Table 3**     Cross-validation results for monosource learning on data coming from the lead I of an ECG and an ABP channel

|  | *Monosource ECG* | | | *Monosource ABP* | | |
|---|---|---|---|---|---|---|
|  | TrAcc | Acc | Comp | TrAcc | Acc | Comp |
| Sr | 0.62 | 0.60 | 7 | 1 | 0.98 | 5 |
| Ves | 0.981 | 0.94 | 5 | 0.990 | 0.76 | 4/5/3 |
| Bige | 1 | 0.98 | 4 | 0.962 | 0.80 | 3 |
| doublet | 1 | 1 | 4 | 0.942 | 0.78 | 4/5 |
| Vt | 1 | 1 | 3 | 0.996 | 0.86 | 6/4 |
| Svt | 1 | 0.98 | 3 | 1 | 1 | 3 |
| Af | 1 | 1 | 2 | 0.998 | 0.86 | ¾ |

**Table 4** Cross-validation results for naive and biased multisource learnings using the monosource learnings of Table 3

|  | *Naive multisource* | | | *Biased multisource* | | |
|---|---|---|---|---|---|---|
|  | *TrAcc* | *Acc* | *Comp* | *TrAcc* | *Acc* | *Comp* |
| Sr | 0.98 | 0.96 | 4 | 1 | 0.98 | 5 |
| Ves | 0.965 | 0.86 | 4 | 0.999 | 0.88 | 5/5 |
| Bige | 0.997 | 0.86 | 3/3 | 1 | 0.98 | 4 |
| Doublet | 0.98 | 0.84 | 5 | 1 | 1 | 4 |
| Vt | 1 | 1 | 3 | 1 | 1 | 3 |
| Svt | 1 | 0.98 | 2 | 1 | 1 | 3 |
| Af | 1 | 1 | 2 | 1 | 1 | 2 |

**Figure 8** **Example of rule learned for class svt from ECG data**

```
class(svt):-
qrs(R0,normal),
p(P1,normal), qrs(R1,normal),
suc(P1,R0), suc(R1,P1),
rr1(R0,R1,short),
p(P2,normal), qrs(R2,normal),
suc(P2,R1), suc(R2,P2),
rythm(R0,R1,R2,regular).
```

**Figure 9** Example of rule learned for class svt from ABP data

```
class(svt) :-cycle_abp(Dias0,_,Sys0,normal),
cycle_abp(Dias1,normal,Sys1,normal),
suc(Dias1,Sys0),
amp_dd(Dias0,Dias1,normal),
ss1(Sys0,Sys1,short),
ds1(Dias1,Sys1,long).
```

**Figure 10** Example of rule learned for class svt by naive multisource learning

```
class(svt):-
qrs(R0,normal),
cycle_abp(Dias0,_,Sys0,normal),
p(P1,normal),qrs(R1,normal),
suci(P1,Sys0),suc(R1,P1),
systole(Sys1),
rr1(R0,R1,short).
```

**Figure 11** Example of rule learned for class svt by biased multisource learning

```
class(svt):-
qrs(R0,normal),
p(P1,normal),qrs(R1,normal),
suc(P1,R0),suc(R1,P1),
rr1(R0,R1,short),
cycle_abp(Dias0,_,Sys0,normal),
suci(Dias0,R1).
```

*5.4   Learning from a less informative database*

The current medical data we are working on are very well known from the cardiologists; so, we have a lot of background information on them. For example, we know which event or which kinds of relations between events are interesting for the learning process, which kind of constraints exists between events occur on the different sources, etc. This knowledge is very useful to create the learning bias and can partly explain partly the very accurate results obtained in the learning experiments above. These good results can also be explained by the small number of examples available for each arrhythmia and the fact that the examples are not corrupted. In this context, it is very difficult to evaluate the usefulness of multiple data sources to improve the learning performances. We have, thus, decided to set ourselves in a more realistic situation where information about the sources is reduced. In this experiment we do not take into account the P waves or the shape of the QRS on the ECG and the diastole on the ABP channel. This experiment makes sense as far as signal processing is concerned since current signal processing algorithms do not detect accurately P waves on the ECG (cf. Portet, 2008) in presence of an arrhythmia. Besides, in our symbolic description of the ABP channel, the diastole is simply the lowest point between two systoles. This specific point is also difficult to detect. Note that cardiologists view the diastole as the time period between two systoles (the moment during which the ventricles fill with blood). The results of this second experiment are given in Table 5. This time again, the biased multisource rules have as good, or better, results than the monosource rules. For arrhythmias *sr* and *bige* the biased method acts like a voting method and learns the same rules with the same accuracy as the best monosource rules (small variations in accuracies come from the cross validation drawback). For arrhythmias *ves, doublet, vt, svt* and *af*, the biased multisource rules are different from both monosource rules corresponding to the same arrhythmia and accuracy and test are better than for the monosource rules.

**Table 5**     Cross-validation results for monosource and biased multisource learnings on data coming from the lead I of an ECG without knowledge on P wave nor QRS shape and from an ABP channel without using information on the diastole

|         | Monosource ECG | | | Monosource ABP | | | Biased multisource | | |
|---------|------|------|------|------|------|------|------|------|------|
|         | *TrAc* | *Acc* | *Comp* | *TrAc* | *Acc* | *Comp* | *TrAc* | *Acc* | *Comp* |
| Sr      | 0.48  | 0.44 | 6   | 1     | 0.98 | 5   | 1     | 1    | 5     |
| Ves     | 0.52  | 0.46 | 6/6 | 0.928 | 0.80 | 5   | 0.942 | 0.76 | 8/6/5 |
| Bige    | 0.98  | 0.90 | 6   | 0.997 | 0.84 | 4/5 | 0.999 | 0.88 | 4/5   |
| Doublet | 0.851 | 0.78 | 5/5 | 0.982 | 0.86 | 4/5 | 0.993 | 0.88 | 4/4   |
| Vt      | 0.883 | 0.72 | 6   | 0.93  | 0.82 | 4/4 | 0.97  | 0.8  | 6/4/7 |
| Svt     | 0.96  | 0.96 | 6   | 0.96  | 0.82 | 4   | 0.99  | 0.96 | 8     |
| Af      | 0.977 | 0.9  | 4/5 | 0.978 | 0.78 | 5/5 | 0.984 | 0.82 | 5/4   |

*5.5   Learning on complementary data*

To emphasise the interest of multiple sources (still in a non noisy context), we have tested multisource learning on truly complementary data. To our knowledge, there is no real annotated multisource database with complementary data referenced in the literature.

Thus, we have chosen to use two virtual sources, the first one gives information only on the P wave of an ECG, the second one gives only information on the QRS complex of the same ECG, without taking into account the shape of the wave. Such data can be found by using an EECG (œsophagal electrocardiogram) and channel V of the ECG. The first results of this study are given in Tables 6 and 7. The results show that the accuracy is really improved by using multiple sources (especially for the naive method) not only during the learning process but also during the test step. Besides, results with the biased multisource learning method are comparable to those obtained with the naive multisource learning on the training accuracy and worse on the test accuracy. However, the efficiency is still a lot better in the biased multisource case (efficiency results are very comparable to those shown in Tables 1 and 2, so they are not repeated here).

**Table 6**     Cross validation results for monosource learning on data coming from a source that only describes QRS complexes and another source that only describes P waves

|         | P wave only | | | QRS complex only | | |
|---------|-------|------|------|-------|------|------|
|         | *TrAcc* | *Acc* | *Comp* | *TrAcc* | *Acc* | *Comp* |
| Sr      | 0.62  | 0.62 | 7    | 0.38  | 0.36 | 5    |
| Ves     | 0.76  | 0.74 | 6    | 0.42  | 0.4  | 5    |
| Bige    | 1     | 0.98 | 4    | 0.96  | 0.92 | 4    |
| Doublet | 0.78  | 0.72 | 3/7  | 0.92  | 0.92 | 4    |
| Vt      | 0.92  | 0.9  | 2    | 0.901 | 0.84 | 5    |
| Vst     | 1     | 1    | 2    | 0.96  | 0.94 | 5    |
| At      | 0     | 0    | 0    | 0.94  | 0.86 | 4/4  |

**Table 7**     Cross validation for naive and biased multisource learnings using monosource rules of Table 6.

|         | Naive multisource | | | Biased multisource | | |
|---------|-------|------|------|-------|------|------|
|         | *TrAcc* | *Acc* | *Comp* | *TrAcc* | *Acc* | *Comp* |
| Sr      | 0.6   | 0.6  | 7    | 0.63  | 0.44 | 11   |
| Ves     | 1     | 0.94 | 6    | 0.98  | 0.8  | 6    |
| Bige    | 1     | 0.98 | 4    | 1     | 0.86 | 4    |
| doublet | 1     | 1    | 4    | 1     | 0.86 | 6    |
| Vt      | 1     | 1    | 4    | 0.92  | 0.72 | 2    |
| Vst     | 1     | 1    | 2    | 1     | 0.86 | 2    |
| Af      | 0.98  | 0.94 | 7    | 0.94  | 0.96 | 4    |

# 6   Related work

There has been an increased interest in multisource learning problems in the past few years (Multiview, 2005); in particular, since the work of Blum and Mitchell on *co-training* (Blum and Mitchel, 1998*)*. Co-training is a semi-supervised learning method that uses several views of a same concept to learn accurate classifiers from a small number of labelled examples and a big number of unlabelled examples. In our application, the concept could be an arrhythmia and the different views could be the different sources. However, the co-training algorithm results in the best classifier that can be learned from the data coming from a single source and so, the learned rules contain only events from one particular source. On the contrary, our aim is to learn composite multisource rules i.e., rules that contain events from all the considered sources and relations between those events.

This work is also related to the work on parallel universes as described by Berthold et al. (Dagstuhl Seminar, 2007) which aims at constructing a global model from a set of connected universes. Parallel universes include work on data fusion, ensemble methods (Diettrich, 2000) or feature selection (Blum and Langley, 1997). Data fusion consists of merging data from different sources for different tasks. The expected benefit of data fusion is a better precision or a better accuracy when data are uncertain. There is a huge amount of work in this area, so we focus on work dealing with symbolic data fusion which is our main concern. In the field of symbolic machine learning, symbolic data fusion can be achieved either at the input level or at the output level. In the first setting, the learning data are merged *a priori* and the multisource learning process is reduced to a monosource one. In the second setting, the learning results are merged *a posteriori* and more or less sophisticated voting schemes such as ensemble methods are used to obtain a multisource solution. The method described in this paper is quite different; the results provided by the monosource classifiers are used to restrict the hypothesis vocabulary, and consequently, the set of hypotheses, of a final global learning step which gives as result a global classifier. Sometimes the process reduces to a voting method i.e., the global classifier is equivalent to one of the monosource classifiers. Sometimes the global classifier is a brand new one establishing new relations between objects from different sources. Part of our process can be seen as a feature selection method since a subset of the initial hypothesis vocabulary used on the different sources is selected.

To the best of our knowledge, this paper is the first attempt to formalise the idea of multisource learning in ILP and to use this new ILP paradigm in a real application. In the application point of view, this work is closely related to the recent work from Syed et al. (2007) which aims at analysing off-line large amount of cardio-vascular data. In this work, they also first transform the raw signals into symbolic strings using clustering methods. This method allows them to detect interesting patterns in the signals without using any background information on the domain. The authors argue that the symbolic transformation allows to drastically reduce the number of information that must be process in order to discover interesting patterns. They then use known algorithms on strings to find repeating periods in the strings, recurrent transient or high entropy patterns. Using their techniques on different signals, they can also detect multimodal trends. The class of cardiac diseases that are tackled by their method is less precise than in our case because patterns are recognised on a cardiac cycle basis and not on a wave basis. Besides, because they do not use any apriori information on the domain (they do not know what are the interesting events to detect), their method is not usable on-line and on a small amount of data. Besides, their paper does not give any quantitative measures on the advantage of using multiple sources compare to only one source. Our work is also related to intelligent monitoring (Coiera, 1993). In this paper, Coiera describes how to build an intelligent monitoring system in four steps: signal processing (i.e., selecting the sensors that gives the most relevant information), signal processing validation (i.e., integrating an intelligent alarm manager which can rule out some signals when they are useless, for example, too noisy), pattern extraction or temporal abstraction (i.e., transforming the signals into symbolic time-stamped objects) and inference engine (i.e., establishing a diagnosis). Our aim is to design an intelligent cardiac monitoring system that efficiently implements the four steps above. The three first steps have been developed in our monitoring system *Calicot* and the ILP method has already been successfully used to learn rules that can be used for the automated diagnostic of cardiac arrhythmias from ECG data (Carrault et al., 2003). However, no studies have been done using ILP on pressure channel data or on multisource data. Existing complete medical intelligent monitoring systems that use multiple sources to give a diagnosis often rely on expert rules (*Guardian*: Larsson and Hayes-Roth, 1998), (*Déjà Vu*: Dojat et al., 1998) or on 'black box' algorithms such as SVM that imply less interpretable results (Morik et al., 2000). In Calicot, arrhythmias are learned as temporal patterns from abstracted signal data. The physicians can then verify the relevance of alerts emitted by the system.

## 7 Discussion and conclusion

We have presented a technique to learn rules from multisource data with an ILP method in order to improve the detection and the recognition of cardiac arrhythmias in a monitoring context. To reduce the computation time of a straightforward multisource learning from aggregated examples, we propose a method to design an efficient bias for multisource learning. This bias is constructed from the results obtained by learning independently from data associated to the different sources. We have shown that this technique provides rules which always have results of better or equal accuracy than monosource rules and that it is much more efficient than a naive multisource learning.

To obtain composite multisource rules, i.e., rules that contain events occurring on all the sources, with the biased multisource method, the learned monosource rules must contain common relational predicates (here, the succession relation). If it is not the case, there is no possible layout between events occurring on the different sources. In the latter case, the biased multisource method behaves as a voting method and learns the best monosource rules. This is also true when data from the different sources are redundant. On the contrary, when the sources are really complementary, the biased multisource method gives very good results compared to monosource learning.

The main drawback of the method is the high number of bottom clauses that can be generated if there are no physiological constraints available to forbid some of the sequences and also if the number of sources increases. To cope with this problem, a first step of frequent pattern mining (Agrawal et al., 1993) can be done directly on the aggregated examples before the learning step to discover automatically the physiological constraints.

In future work, the CALICOT system will be tested with the new multisource rules in order to assess its performance in a realistic clinical noisy context. The learning method will be tested on corrupted data. Besides, the recognition and the diagnosis of cardiac arrhythmias will be more deeply evaluated by experts.

## Acknowledgements

During this work, Elisa Fromont was supported by the French National Network for Health Technologies as a member of the CEPICA project. This project is in collaboration with LTSI-Université de Rennes1, ELA-Medical and Rennes University Hospital.

## Notes

[1]ICL is still developed in the CLASSIC'CL (Stolle et al., 2005) and in the ACE (http://www.cs.kuleuven.be/~dtai/ACE/) systems.

[2]The accuracy is defined by the formula (TP + TN/(TP + TN + FN + FP)) where True Positive (TP) is the number of positive examples classified as true, True Negative (TN) the number of negative examples classified as false, False Negative (FN) is the number of positive examples classified as false and False Positive (FP), the number of negative examples classified as true.